\documentclass[10pt,twocolumn,letterpaper]{article}

\usepackage{cvpr}              
\usepackage[accsupp]{axessibility}
\usepackage{graphicx}
\usepackage{amsmath}
\usepackage{amssymb}
\usepackage{booktabs}

\usepackage[pagebackref,breaklinks,colorlinks]{hyperref}

\usepackage[capitalize]{cleveref}
\crefname{section}{Sec.}{Secs.}
\Crefname{section}{Section}{Sections}
\Crefname{table}{Table}{Tables}
\crefname{table}{Tab.}{Tabs.}

\def\cvprPaperID{3404} 
\def\confName{CVPR}
\def\confYear{2023}

\begin{document}
	
	\title{Low-Light Image Enhancement via Structure Modeling and Guidance}
	
\author{
 Xiaogang Xu$^{1}$ \quad Ruixing Wang$^{2}$ \quad Jiangbo Lu$^{3}$$^*$ \\
$^1$ Zhejiang Lab \quad $^2$ Honor Device Co., Ltd. \quad $^3$ SmartMore Corporation\\
{\tt \small xgxu@zhejianglab.com, ruixingw@hustunique.com, jiangbo@smartmore.com}
}
\maketitle

\renewcommand{\thefootnote}{\fnsymbol{footnote}} 
\footnotetext[1]{Corresponding author.}

\begin{abstract}
This paper proposes a new framework for low-light image enhancement by simultaneously conducting the appearance as well as structure modeling. It employs the structural feature to guide the appearance enhancement, leading to sharp and realistic results. The structure modeling in our framework is implemented as the edge detection in low-light images. It is achieved with a modified generative model via designing a structure-aware feature extractor and generator. The detected edge maps can accurately emphasize the essential structural information, and the edge prediction is robust towards the noises in dark areas. Moreover, to improve the appearance modeling, which is implemented with a simple U-Net, a novel structure-guided enhancement module is proposed with structure-guided feature synthesis layers. The appearance modeling, edge detector, and enhancement module can be trained end-to-end. The experiments are conducted on representative datasets (sRGB and RAW domains), showing that our model consistently achieves SOTA performance on all datasets with the same architecture.
The code is available at \href{https://github.com/xiaogang00/SMG-LLIE}{https://github.com/xiaogang00/SMG-LLIE}.
\end{abstract}

\section{Introduction}
\begin{figure}[t]
	\centering
	\captionsetup[subfigure]{labelformat=empty}
	\begin{subfigure}[c]{0.15\textwidth}
		\centering
		\includegraphics[width=1.01in]{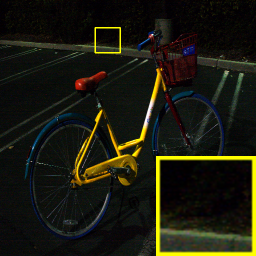}
		\vspace{-1.5em}
		\caption{(a) Input}
	\end{subfigure}
	\begin{subfigure}[c]{0.15\textwidth}
		\centering
		\includegraphics[width=1.01in]{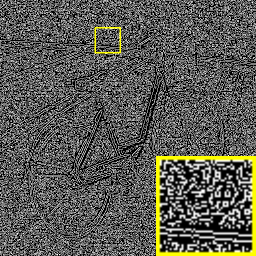}
		\vspace{-1.5em}
		\caption{(b) Structure of (a)}
	\end{subfigure}
	\begin{subfigure}[c]{0.15\textwidth}
		\centering
		\includegraphics[width=1.0in]{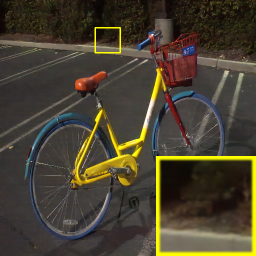}
		\vspace{-1.5em}
		\caption{(c) SNR (CVPR 2022)}
	\end{subfigure}
	\vspace{0.1em}
	\begin{subfigure}[c]{0.15\textwidth}
		\centering
		\includegraphics[width=1.0in]{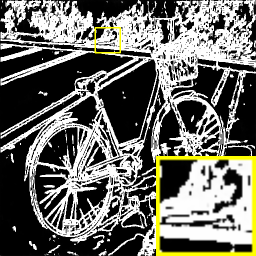}
		\vspace{-1.5em}
		\caption{(d) Structure Modeling}
	\end{subfigure}
	\begin{subfigure}[c]{0.15\textwidth}
		\centering
		\includegraphics[width=1.0in]{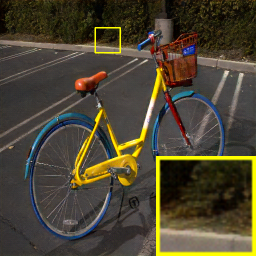}
		\vspace{-1.5em}
		\caption{(e) Ours}
	\end{subfigure}
	\begin{subfigure}[c]{0.15\textwidth}
		\centering
		\includegraphics[width=1.0in]{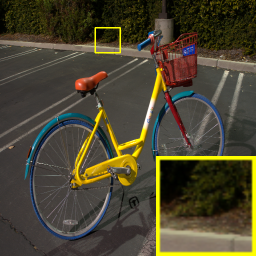}
		\vspace{-1.5em}
		\caption{(f) Ground Truth}
	\end{subfigure}
	\vspace{-0.1in}
	\caption{A challenging low-light frame (a), from SID-sRGB~\cite{chen2018learning}, enhanced by a SOTA method (c) and our method (e). Our method can synthesize the structure map (d) from the input image, leading to clearer details, more distinct contrast, and more vivid color. Although (c) has high PSNR as 28.17, its SSIM is low as 0.75. Ours achieves high scores for both dB and SSIM, as 28.60dB and 0.80.
	}
	\vspace{-0.15in}
	\label{fig:teaser}
\end{figure}

The low-light enhancement aims to recover normal-light and noise-free images from dark and noisy pictures, which is a long-standing and significant computer vision topic.
It has broad application fields, including low-light imaging~\cite{wang2019low,lu2022progressive,hao2020low}, and also benefits many downstream vision tasks, e.g., nighttime detection~\cite{wang2021hla,yu2021single,ma2022toward}.
Some methods have been proposed to tackle the low-light enhancement problem. They design networks that learn to manipulate color, tone, and contrast~\cite{wang2013naturalness,fu2016weighted,ying2017new,guo2016lime}, and some recent works also account for noise in images~\cite{liu2021retinex,xu2020learning}.
Most of these works optimize the appearance distance between the output and the ground truth. However, they ignore the explicit modeling of structural details in dark areas and thus resulting in blurry outcomes and low SSIM~\cite{wang2004image} values, as shown in Fig.~\ref{fig:teaser}.
Some works~\cite{zhu2020eemefn,rana2021edge} have noticed the effect of employing structural information, e.g., edge, to promote the enhancement. Edge guides the enhancement by distinguishing between different parts in the dark regions.
Moreover, adding sensible edge priors into dark regions reduces the ill-posed degree in optimizing the appearance reconstruction.
These frameworks~\cite{zhu2020eemefn,rana2021edge} perform the structure modeling with encoder-decoder-based networks and a regression loss.
However, the corresponding structure modeling results are not satisfying due to the uncertainty in the dark areas caused by severely poor visibility and noise. 
Furthermore, the strategy of using the extracted structural information needs to be improved from the existing straightforward concatenation approach~\cite{zhu2020eemefn,rana2021edge}.

\begin{figure}[t]
	\begin{center} 
		\includegraphics[width=1.0\linewidth]{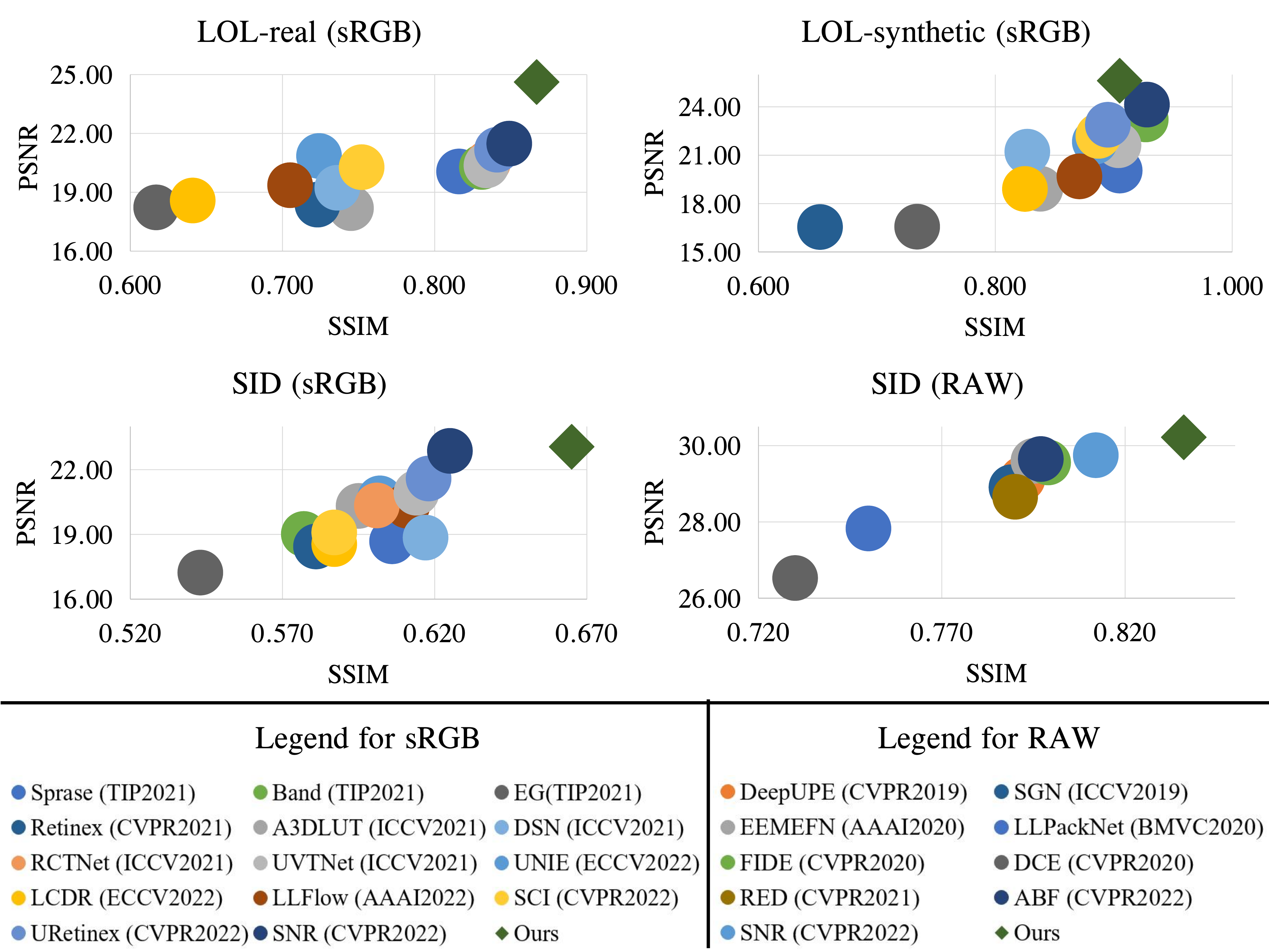}
	\end{center}
	\vspace{-0.25in}
	\caption{Our method {\em consistently\/} achieves SOTA performance on {\em different sRGB/RAW datasets\/} with the same network architecture. 
	}
	\vspace{-0.25in}
	\label{fig:teaser-com}
\end{figure}

In this paper, we propose to utilize a generative model $\mathcal{S}$ trained with a GAN loss to perform the structure modeling with the form of edges. Then, we design a new mechanism $\mathcal{E}$ to facilitate the initial low-light appearance enhancement (the module is denoted as $\mathcal{A}$) with structure-guided feature synthesis.
With effective structure modeling and guidance, our framework can output sharp and realistic results with satisfactory reconstruction quality as shown in Fig.~\ref{fig:teaser}.

Compared with previous structure modeling networks, the proposed generative model $\mathcal{S}$ has two significant modifications. First, we notice the impact of providing structure-aware descriptors into both the encoder and decoder of $\mathcal{S}$, disentangling the appearance representation and underlining the structural information.
Thus, we design a Structure-Aware Feature Extractor (SAFE) as the encoder part, which extracts structure-aware features from the dark image and its gradients via spatially-varying operations (achieved with adaptive long-range and short-range computations).
The extracted structure-aware tensors are then fed into the decoder part to generate the desired structure maps.
Moreover, different from current approaches, which employ the structure maps of normal-light images to conduct the regression learning, we find the nice property of using a GAN loss. The GAN loss can reduce the artifacts in the generated structure maps that are caused by the noise and invisibility, highlighting the essential structure required for enhancement.
The backbone of $\mathcal{S}$ is implemented as a modified StyleGAN.

To boost the appearance by leveraging the obtained structure maps, we design a Structure-Guided Enhancement Module (SGEM) as $\mathcal{E}$.
The main target of SGEM is to learn the residual, which can improve the initial appearance modeling results.
In SGEM, spatially-adaptive kernels and normalization parameters are generated according to the structure maps.
Then, the features in each layer of the SGEM's decoder will be processed with spatially-adaptive convolutions and normalization. 
Although the overall architecture of SGEM takes the form of a simple U-Net~\cite{ronneberger2015u}, it can effectively enhance the original appearance.

$\mathcal{S}$, $\mathcal{A}$, and $\mathcal{E}$ can be trained end-to-end simultaneously.
Extensive experiments are conducted on representative benchmarks.
Experimental results show that our framework achieves SOTA performance on both PSNR and SSIM metrics with the same architecture on all datasets, as shown in Fig.~\ref{fig:teaser-com}.
In summary, our work's contribution is four-fold.
\begin{enumerate}
	\item We propose a new framework for low-light enhancement by conducting structure modeling and guidance simultaneously to boost the appearance enhancement.
	\item We design a novel structure modeling method, where structure-aware features are formulated and trained with a GAN loss.
	\item A novel structure-guided enhancement approach is proposed for appearance improvement guided by the restored structure maps.
	\item Extensive experiments are conducted on different datasets in both sRGB and RAW domains, showing the effectiveness and generalization of our framework.
\end{enumerate}

\section{Related Work}

\noindent\textbf{Low-light enhancement with learning.}
The current low-light image enhancement methods with deep learning mainly optimize the appearance reconstruction error between the output and the ground truth~\cite{yan2014learning, yan2016automatic, lore2017llnet, cai2018learning,Zero-DCE,zamir2020learning,xu2020learning,zeng2020learning,kim2021representative,zhao2021deep,zheng2021adaptive,wang2021real,liu2021retinex,yang2021sparse,jiang2021enlightengan,yang2021band}.
However, the enhanced appearance tends to be blurry in the dark regions even with SOTA approaches~\cite{xu2022snr}.
To this, some works utilize the structural information~\cite{zhu2020eemefn,kim2021deep} to enhance appearance. 
To obtain the structure maps, some of them use offline-computed edges/gradients~\cite{kim2021deep,ren2019low,hao2020low,tanaka2019gradient} or offline-trained networks~\cite{liang2022semantically} which are not adaptively optimized with the low-light data and result in artifacts inevitably.
Although several methods~\cite{zhu2020eemefn,rana2021edge} have been proposed to train an edge detector and an image enhancement network simultaneously with both regression losses, their improvements are still limited.
Moreover, existing strategies mainly concatenate the extracted structure maps with the images to facilitate the enhancement~\cite{kim2021deep,zhu2020eemefn,rana2021edge}, which, however, can not set the guidance in all layers of the enhancement network.
Different from current works, we propose a new generative model to robustly extract the edge map from a dark input image with the GAN loss, providing the critical edge information for enhancement in every feature layer.

\begin{figure*}[t]
	\begin{center}
		\includegraphics[width=1.0\linewidth]{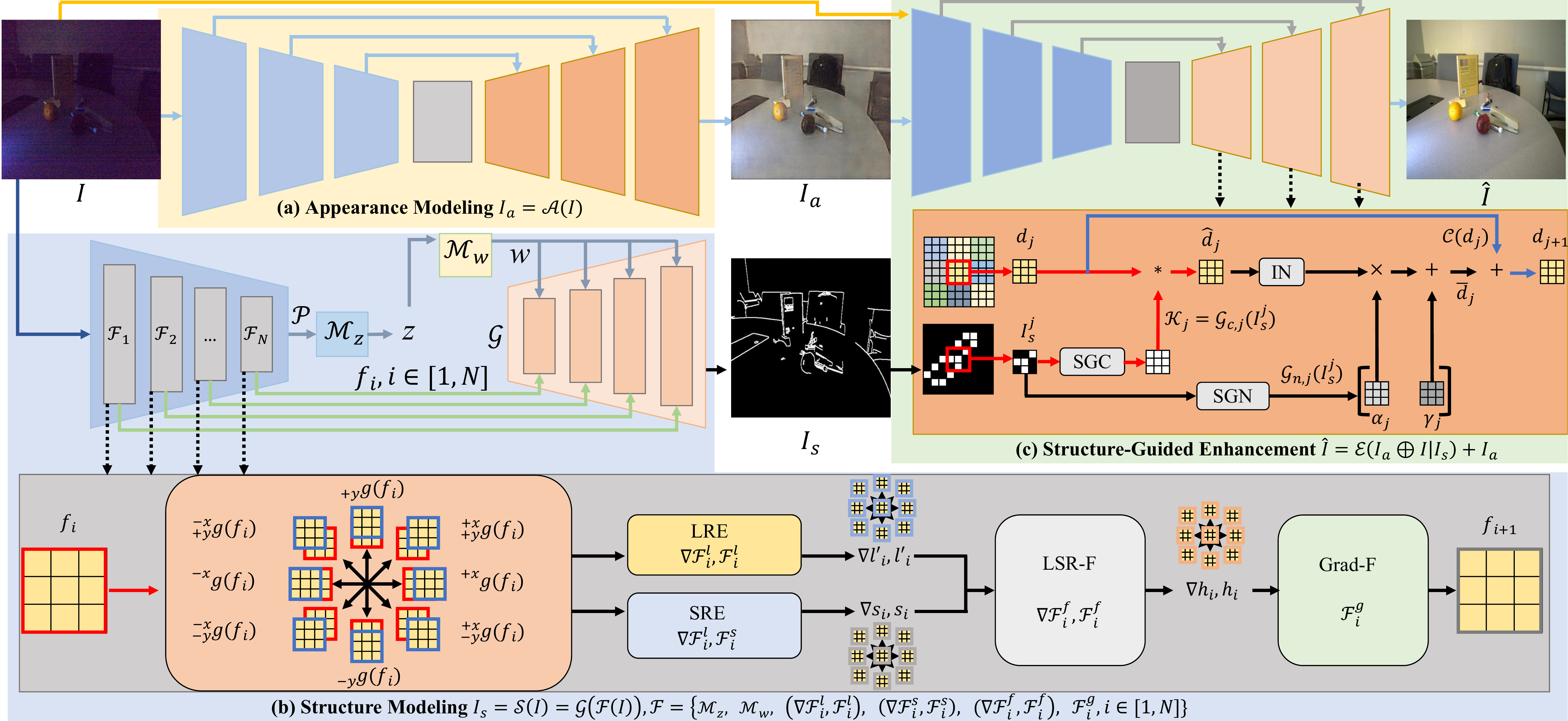}
	\end{center}
	\vspace{-0.2in}
	\caption{The overview of our framework with explicit appearance modeling $\mathcal{A}$, structure modeling $\mathcal{S}$, and SGEM $\mathcal{E}$. The supervision for $I_a$ and $\widehat{I}$ is the normal-light image $\bar{I}$, and for $I_s$ is the edge $\bar{I}_s$ extracted from $\bar{I}$. The overall framework can be trained end-to-end.
	}
	\label{fig:framework}
	\vspace{-0.2in}
\end{figure*}

\noindent\textbf{Generative model for restoration.}
The recent development on the generative model enables a wide range of networks, e.g., StyleGAN~\cite{karras2020analyzing}, to achieve significant restoration effects in some tasks, e.g., face restoration~\cite{yang2021gan,he2022gcfsr,zhu2022blind}.
However, their performances are highly dependent on the pre-trained model within several categories.
In this paper, we demonstrate that generative models with the GAN loss can be utilized in a novel way to synthesize structural information for the restoration task of low-light image enhancement.
Such an approach does not rely on the pre-trained model, and such an idea can actually be extended to various restoration tasks, opening up a new research direction.

\section{Method}
The overview of our framework can be viewed in Fig.~\ref{fig:framework}. 
The proposed framework conducts explicit modeling for both appearance $\mathcal{A}$ as well as structure $\mathcal{S}$, and they will be described in Sec.~\ref{sec:appear} and Sec.~\ref{sec:structure}, respectively.
The structure maps are employed to guide the enhancement of appearance modeling with the module of $\mathcal{E}$, which will be introduced in Sec.~\ref{sec:enhancement}.
The overall framework can be trained end-to-end as elaborated in Sec.~\ref{sec:loss}.

\vspace{-0.05in}
\subsection{Appearance Modeling}
\label{sec:appear}
In our framework, given an input low-light image $I$, we adopt a common U-Net~\cite{chen2018learning} as the network for appearance modeling, as shown in Fig.~\ref{fig:framework} (a), where $I_a=\mathcal{A}(I)$.
The experimental results show that our approach combined with the structural modeling can achieve SOTA performance, with such an ordinary network for appearance prediction.

\vspace{-0.05in}
\subsection{Structure Modeling}
\label{sec:structure}
In this paper, we adopt edge information for structure modeling, which is a general representation applicable to various scenes.
We find that the effective generative model, e.g., the StyleGAN~\cite{karras2020analyzing}, can be modified to be a strong structure estimator $\mathcal{S}$ for low-light images as shown in Fig.~\ref{fig:framework} (b). Equipped with both the structure regression loss and the GAN loss, $\mathcal{S}$ can accurately predict the effective structure maps that are beneficial for appearance enhancement. 
In this paper, we use the backbone of StyleGAN to formulate $\mathcal{S}$.
Given the input image $I$, we can obtain its structure map $\mathcal{S}$ as
\begin{equation}
	\small
	I_s=\mathcal{S}(I)=\mathcal{G}(\mathcal{F}(I)),
\end{equation}
where $\mathcal{F}$ is the encoder part, as ``Structure-Aware Feature Extractor (SAFE)" that contains $N$ layers and two mapping functions, and $\mathcal{G}$ is decoder part, as ``Structure-Aware Generator (SAG)". $\mathcal{F}$ and $\mathcal{G}$ enable our framework to perform better structure modeling than previous generative models and edge detectors.

\vspace{-0.05in}
\subsubsection{Structure-Aware Feature Extractor (SAFE)}
Although the traditional encoder-based generative model can be employed as the structure extractor, its performance can be improved by feeding structural features into both the encoder and the generator. 
Previous encoder-based generative models extract information from only the content feature, which is suitable for the content generation (e.g., sRGB images) while insufficient for structure modeling.
Moreover, the conventional feature extractor of the generative model mainly utilizes the short-range operation~\cite{tov2021designing,richardson2021encoding,wang2022high}, e.g., CNN. However, as indicated in~\cite{xu2022snr}, formulating useful representations in dark areas with low Signal-to-Noise Ratio (SNR) needs long-range operations.

Thus, we design the SAFE in $S$ as shown in Fig.~\ref{fig:framework} (b), whose main target is to extract the reliable structural features that can be utilized next in the SAG.
Different from conventional generative models' encoders, SAFE obtains the required information from both content and gradient maps which are helpful for edge modeling. Spatially-varying operations are formulated in SAFE for feature extraction.

\noindent\textbf{The formulation of gradient maps.} As shown in Fig.~\ref{fig:framework} (b), the input image $I$ is extracted by a multi-layer encoder, $\mathcal{F}_1, ..., \mathcal{F}_N$, with down-sampling in each layer.
Suppose the input feature of each layer is $f_i, i\in[1,N]$, we obtain the gradient map of $f_i$ in the directions of $x$- and $y$-axis, as 
$\{ g_{+x}(f_i)$, $g_{-x}(f_i)$, $g_{+y}(f_i)$, $g_{-y}(f_i)$, $g_{+x, +y}(f_i)$, $g_{+x, -y}(f_i)$, $g_{-x, +y}(f_i)$, $g_{-x, -y}(f_i) \}$, where $g$ is the function to compute the first order gradient.
These gradient maps emphasize the edge regions, benefiting the formulation of the structural features.

\noindent\textbf{Spatially-varying feature extraction for structure.}
We design the spatially-varying operations in the encoder to effectively extract structural features.
For $f_i$ and $g(f_i)$, we set the corresponding Long-Range Encoder (LRE) and Short-Range Encoder (SRE) modules.
The transformer blocks are proven to have extraordinary ability in the long-range modeling~\cite{yuan2021tokens,wu2021cvt,yuan2021incorporating,liu2021swin,chen2021pre,wang2021uformer}, and the CNN blocks are capable of extracting short-range characteristics.
Thus, LRE and SRE are implemented as the transformer and CNN blocks, respectively. 
To build an efficient transformer module, we adopt the transformer block with the window-based attention mechanism~\cite{liu2021swin} and local-enhanced forward module~\cite{wang2021uformer}.
Suppose LRE and SRE modules are denoted as $\mathcal{F}_l$ and $\mathcal{F}_s$, the feature extraction process can be written as
\begin{equation}
	\small
	\begin{aligned}
		&l_i=\quad \mathcal{F}_i^l(f_i), \quad \quad \quad \; \; \; \; s_i=\quad \mathcal{F}_i^s(f_i),\\
		\bigtriangledown &l_{i}=\bigtriangledown\mathcal{F}_i^l(\bigtriangledown g(f_i)), \quad
		\bigtriangledown s_{i}=\bigtriangledown \mathcal{F}_i^s(\bigtriangledown g(f_i)),
	\end{aligned}
\end{equation}
where $\bigtriangledown \in \{ ^{+x}, ^{-x}, _{+y}, _{-y}, _{+y}^{+x}, _{+y}^{-x}, _{-y}^{+x}, _{-y}^{-x}\}$,
$l_i$ is the long-range feature and $s_i$ is the short-range feature, respectively.
We adopt another Long-Short-Range Fusion (LSR-F) module to merge the outputs from long-range and short-range operations, whose architecture is the multilayer perceptron.
Suppose LSR-F is denoted as $\mathcal{F}_i^{f}$, then the spatially-varing extracted features for $f$ and $g$ can be formulated as
\begin{equation}
	\small
	\begin{aligned}
		&\quad h_i=\mathcal{F}_i^{f}(l_i, s_i),&\bigtriangledown h_{i}=\bigtriangledown \mathcal{F}_i^{f}(\bigtriangledown l_i, \bigtriangledown s_i).
	\end{aligned}
\end{equation}
Moreover, these spatially-varying extracted features for different directions can be adaptively merged with the Gradient Fusion (Grad-F) module, as 
\begin{equation}
	\small
	\begin{aligned}
		f_{i+1}=\mathcal{F}_{i}^g(h_i, \, &^{+x}h_{i},\, ^{-x}h_{i},\, _{-y}h_{i}, \,_{+y}h_{i}, \\ &_{+y}^{+x}h_{i},\, _{+y}^{-x}h_{i}, \,_{-y}^{+x}h_{i}, \,_{-y}^{-x}h_{i}),
	\end{aligned}
\end{equation}
where $\mathcal{F}_{i}^g$ denotes the Grad-F module.

\vspace{-0.05in}
\subsubsection{Structure-Aware StyleGAN Generator (SAG)}
After obtaining the structure-aware features $f_i, i\in[1,N]$, the next step is to get the $w$ space of the StyleGAN backbone. Following previous StyleGAN models with the encoder for the degraded data~\cite{yang2021gan}, $f_N$ is first mapped into the $z$ space and then mapped to the $w$ space, as
\begin{equation}
	\small
	w=\mathcal{M}_w(z)=\mathcal{M}_w(\mathcal{M}_z(\mathcal{P}(f_N))),
\end{equation}
where $\mathcal{M}_w$ and $\mathcal{M}_z$ are the mapping functions and $\mathcal{P}$ is the pooling operation.
The extracted $f_i, i\in[1,N]$ serve as the noise map to feed the structural information into the generator $\mathcal{G}$, as shown in Fig.~\ref{fig:framework} (b).

\vspace{-0.05in}
\subsection{Structure-Guided Enhancement Module}
\label{sec:enhancement}
\vspace{-0.05in}
The structure modeling of the input low-light image $I$, as $I_s$, can be utilized to enhance the appearance predictions $I_a$.
The enhancement can be interpreted from two aspects. First, using the structural information can enhance the image details, especially for the dark areas with low SNR, helping the generation of sharp and realistic results.
The second effect is that the edge information help to distinguish different dark areas and build better relations among them.

Suppose the Structure-Guided Enhancement Module (SGEM) is denoted as $\mathcal{E}$, which can also be implemented as a simple U-Net.
As shown in Fig.~\ref{fig:framework} (c), its input is the concatenation of $I_a$ and $I$ whose difference can provide the coarse direction for the enhancement on $I_a$.
In this way, the enhancement can be denoted as
\begin{equation}
	\small
	\widehat{I}=\mathcal{E}(I_a\oplus I | I_s)+I_a,
\end{equation}
where $\oplus$ is the concatenation operation, and $I_s$ is the condition for the enhancement that is achieved by ``Structure-Guided Feature Synthesis" in $\mathcal{E}$.

\begin{table*}[t]
	\centering
	\Huge
	\resizebox{1.0\linewidth}{!}{
		\begin{tabular}{c|cccccccccccc}
			\toprule[1pt]
			Methods &SID~\cite{chen2018learning}  & DeepUPE~\cite{wang2019underexposed} &KIND~\cite{zhang2019kindling} & DeepLPF~\cite{moran2020deeplpf} & FIDE~\cite{xu2020learning}&LPNet~\cite{li2020luminance}&MIR-Net~\cite{zamir2020learning}&RF~\cite{kosugi2020unpaired} &3DLUT~\cite{zeng2020learning}&UNIE~\cite{jin2022unsupervised} &LCDR~\cite{wang2022local}&LLFlow~\cite{wang2022low} \\
			PSNR &13.24&13.27&14.74&14.10 &16.85&17.80 &20.02&14.05&17.59 &20.85 &18.57 &19.36  \\
			SSIM&0.442&0.452&0.641&0.480 &0.678&0.792 &0.820&0.458& 0.721 &0.724 &0.641 &0.705  \\
			\hline
			Methods &A3DLUT~\cite{wang2021real} &Band~\cite{yang2021band}&EG~\cite{jiang2021enlightengan}&Retinex~\cite{liu2021retinex}&Sparse~\cite{yang2021sparse} & DSN~\cite{zhao2021deep}&RCTNet~\cite{kim2021representative}&UTVNet~\cite{zheng2021adaptive}&SCI~\cite{ma2022toward}&URetinex~\cite{wu2022uretinex}&SNR~\cite{xu2022snr}&Ours \\
			PSNR&18.19 & 20.29&18.23&18.37&20.06&19.23&20.51&20.37&20.28&21.16&21.48&\textbf{24.62}\\
			SSIM&0.745&0.831 &0.617&0.723&0.815&0.736&0.831&0.834&0.752&0.840&0.849&\textbf{0.867}\\
			\bottomrule[1pt]
	\end{tabular}}
	\vspace*{-0.15in}
	\caption{Quantitative comparison on the LOL-real dataset.}
	\label{comparison3}
	
	\vspace*{0.05in}
	\resizebox{1.0\linewidth}{!}{
		\begin{tabular}{c|cccccccccccc}
			\toprule[1pt]
			Methods &SID~\cite{chen2018learning}  & DeepUPE~\cite{wang2019underexposed} &KIND~\cite{zhang2019kindling} & DeepLPF~\cite{moran2020deeplpf} & FIDE~\cite{xu2020learning}&LPNet~\cite{li2020luminance}&MIR-Net~\cite{zamir2020learning}&RF~\cite{kosugi2020unpaired} &3DLUT~\cite{zeng2020learning}&UNIE~\cite{jin2022unsupervised} &LCDR~\cite{wang2022local}&LLFlow~\cite{wang2022low} \\
			PSNR &15.04&15.08&13.29&16.02 &15.20&19.51 &21.94&15.97& 18.04&21.84& 18.91&19.69  \\
			SSIM&0.610&0.623&0.578&0.587 &0.612& 0.846&0.876&0.632&0.800 & 0.884&0.825&0.871  \\
			\hline
			Methods&A3DLUT~\cite{wang2021real} &Band~\cite{yang2021band}&EG~\cite{jiang2021enlightengan}&Retinex~\cite{liu2021retinex}&Sparse~\cite{yang2021sparse} & DSN~\cite{zhao2021deep}&RCTNet~\cite{kim2021representative}&UTVNet~\cite{zheng2021adaptive}&SCI~\cite{ma2022toward}&URetinex~\cite{wu2022uretinex}&SNR~\cite{xu2022snr}&Ours \\
			PSNR&18.92 & 23.22&16.57&16.55&22.05&21.22&22.44&21.62&22.20&22.89&24.14&\textbf{25.62}\\
			SSIM&0.838 &0.927 &0.734&0.652&0.905&0.827&0.891&0.904&0.887&0.895&\textbf{0.928}&0.905\\
			\bottomrule[1pt]
	\end{tabular}}
	\vspace{-0.15in}
	\caption{Quantitative comparison on the LOL-synthetic dataset.}
	\label{comparison4}
	
	\vspace*{0.05in}
	\resizebox{1.0\linewidth}{!}{
		\begin{tabular}{c|cccccccccccc}
			\toprule[1pt]
			Methods &SID~\cite{chen2018learning}  & DeepUPE~\cite{wang2019underexposed} &KIND~\cite{zhang2019kindling} & DeepLPF~\cite{moran2020deeplpf} & FIDE~\cite{xu2020learning}&LPNet~\cite{li2020luminance}&MIR-Net~\cite{zamir2020learning}&RF~\cite{kosugi2020unpaired} &3DLUT~\cite{zeng2020learning}&UNIE~\cite{jin2022unsupervised} &LCDR~\cite{wang2022local}&LLFlow~\cite{wang2022low} \\
			PSNR &16.97&17.01&18.02&18.07 &18.34&20.08 &20.84&16.44& 20.11&20.67 & 18.55& 20.33  \\
			SSIM&0.591&0.604&0.583&0.600 &0.578&0.598&0.605&0.596 &0.592 &0.602 & 0.587& 0.611 \\
			\hline
			Methods &A3DLUT~\cite{wang2021real} &Band~\cite{yang2021band}&EG~\cite{jiang2021enlightengan}&Retinex~\cite{liu2021retinex}&Sparse~\cite{yang2021sparse} & DSN~\cite{zhao2021deep}&RCTNet~\cite{kim2021representative}&UTVNet~\cite{zheng2021adaptive}&SCI~\cite{ma2022toward}&URetinex~\cite{wu2022uretinex}&SNR~\cite{xu2022snr}&Ours \\
			PSNR&20.32 & 19.02&17.23&18.44&18.68&18.85&20.34&20.93&19.09&21.56&22.87&\textbf{23.18}\\
			SSIM&0.595 &0.577&0.543&0.581&0.606&0.617&0.601&0.614&0.585&0.619&0.625&\textbf{0.664}\\
			\bottomrule[1pt]
	\end{tabular}}
	\vspace{-0.15in}
	\caption{Quantitative comparison on the SID dataset (sRGB domain).}\vspace{-0.2in}
	\label{comparison5}
\end{table*}

\vspace{-0.1in}
\subsubsection{Structure-Guided Feature Synthesis}
Suppose there are $K$ layers in the SGEM's decoder, and the input of each layer is $d_j\in \mathbb{R}^{b\times p \times q}, j\in[1,K]$, where $b$, $p$, and $q$ are channel number, feature height, and width, respectively.
In the $j$-th layer, we first resize the structure map $I_s$ into the same size of $d_j$, as $I^j_s$. To utilize the guidance of the structure map, we propose to generate spatially-varying kernels and normalization maps from $I^j_s$. The generation can be completed with Structure Guided Convolutions (SGC) and Structure Guided Normalizations (SGN), respectively.
As shown in Fig.~\ref{fig:framework} (c), SGC produces different kernels for different locations. The feature map can be processed as
\begin{equation}
	\small
	\mathcal{K}_j=\mathcal{G}_{c,j}(I^j_s),\quad \widehat{d}_j=d_j * \mathcal{K}_j,
\end{equation}
where $\widehat{d}_j\in \mathbb{R}^{b\times p \times q}$, ``$*$" is the convolution operation, $\mathcal{K}_j \in \mathbb{R}^{(b\times (k_h\times k_w))\times p \times q}$ and $\mathcal{G}_{c,j}$ are the synthesized kernels ($k_h$ and $k_w$ are the kernel height and width) and the SGC in the $j$-th layer, respectively.
SGN can predict different normalization maps for different locations to process the feature map, as
\begin{equation}
	\small
	\alpha_j, \gamma_j=\mathcal{G}_{n,j}(I^j_s), \quad \bar{d}_j=IN(\widehat{d}_j)\circ \alpha_j + \gamma_j,
\end{equation}
where $\alpha_j\in \mathbb{R}^{b\times p\times q}$ and $\gamma_j \in \mathbb{R}^{b\times p\times q}$ are the normalization parameters in the $j$-th layer, $\circ$ denotes the Hadamard product, $\bar{d}_j \in \mathbb{R}^{b\times p\times q}$, $IN$ is the instance normalization operation, and $\mathcal{G}_{n,j}$ is the SGN in the $j$-th layer.
In the $j$-th layer, the feature can be enhanced as
\begin{equation}
	\small
	d_{j+1}=\mathcal{C}(d_j) + \bar{d}_j,
\end{equation}
where $\mathcal{C}$ is the original convolutions in U-Net's $j$-th layer.

\subsection{Loss Function}
\label{sec:loss}
Our framework is trained end-to-end, and the loss function can be divided into three parts.

\noindent\textbf{Loss for appearance modeling.}
The loss function for the appearance modeling part $\mathcal{A}$ is the reconstruction error between $I_a$ and the ground truth $\bar{I}$.
The loss is computed at both the pixel level and perceptual level, as
\begin{equation}
	\small
	\mathcal{L}_a=\lVert I_a-\bar{I} \rVert+\lVert \Phi(I_a)-\Phi(\bar{I}) \rVert,
\end{equation}
where $\Phi()$ extracts features from the VGG network~\cite{simonyan2014very}.

\noindent\textbf{Loss for structure modeling.} 
One approach for supervising $\mathcal{S}$ is employing the structure of the normal-light data as the ground truth and utilizing the regression loss. This strategy is adopted by existing edge prediction methods~\cite{pu2022edter,su2021pixel}.
For the input of $I$, we assume the ground truth is the edge map extracted from $\bar{I}$ (using the Canny edge detector~\cite{canny1986computational}), as $\bar{I}_s$.
The regression loss is the binary cross-entropy, as
\begin{equation}
	\small
	\mathcal{L}_s=-[\bar{I}_s \log I_s+ (1-\bar{I}_s) \log (1-I_s)], \; \bar{I}_s=C(\bar{I}),
\end{equation}
where $C$ is the Canny detector, 
Moreover, the invisibility of content and the influence of noise in dark images would aggravate the ill-posed degree of estimating structures. Thus, it is difficult to detect structural details from the dark input picture with only the regression loss.
We find the GAN loss is effective in dealing with the drawback of the regression loss.
The GAN loss is conducted by setting a discriminator $\mathcal{D}$, as
\begin{equation}
	\small
	\begin{aligned}
		\mathcal{L}_g=&\mathbb{E}_{I}(\log(1+\exp(-\mathcal{D}(I_s)))),\\
		\mathcal{L}_d=&\mathbb{E}_{I}(\log(1+\exp(-\mathcal{D}(\bar{I}_s))))+\\
		&\mathbb{E}_{I}(\log(1+\exp(+\mathcal{D}({I}_s)))),\\
	\end{aligned}
\end{equation}
where $\mathcal{L}_{g}$ and $\mathcal{L}_{d}$ are the loss functions for $\mathcal{S}$ and $\mathcal{D}$, respectively, and $\mathbb{E}$ is the mean operation.

\noindent\textbf{Loss for SGEM.}
SGEM $\mathcal{E}$ is to refine the appearance modeling, and the loss is also the reconstruction error, as
\begin{equation}
	\mathcal{L}_m=\lVert \widehat{I}-\bar{I} \rVert+\lVert \Phi(\widehat{I})-\Phi(\bar{I}) \rVert.
\end{equation}

\noindent\textbf{Overall loss function.}
The overall framework is trained in end-to-end manner and the overall loss function is 
\begin{equation}
	\mathcal{L}=\lambda_1 \mathcal{L}_a+\lambda_2 \mathcal{L}_s+\lambda_3 \mathcal{L}_g+\lambda_4 \mathcal{L}_m,
\end{equation}
where $\lambda_1$, $\lambda_2$, $\lambda_3$, $\lambda_4$ are the loss weights.

\section{Experiments}

\begin{figure*}[t]
	\centering
	\begin{subfigure}[c]{0.162\textwidth}
		\centering
		\includegraphics[width=1.12in]{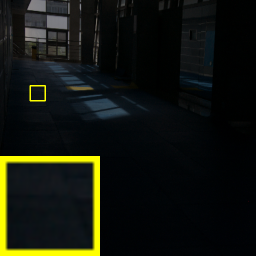}
		\vspace{-1.5em}
		\caption*{Input}
	\end{subfigure}
	\begin{subfigure}[c]{0.162\textwidth}
		\centering
		\includegraphics[width=1.12in]{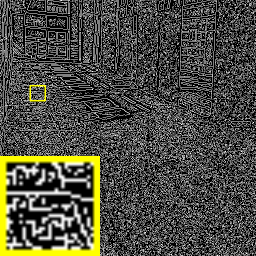}
		\vspace{-1.5em}
		\caption*{Structure of Input}
	\end{subfigure}
	\begin{subfigure}[c]{0.162\textwidth}
		\centering
		\includegraphics[width=1.12in]{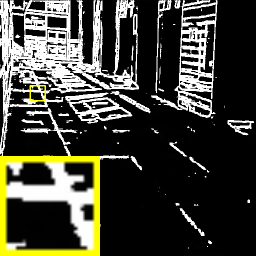}
		\vspace{-1.5em}
		\caption*{Our Structure Map}
	\end{subfigure}
	\begin{subfigure}[c]{0.162\textwidth}
		\centering
		\includegraphics[width=1.12in]{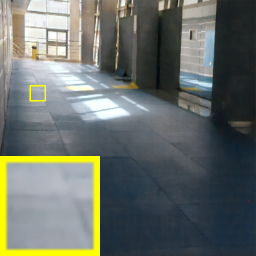}
		\vspace{-1.5em}
		\caption*{SNR~\cite{xu2022snr}}
	\end{subfigure}
	\begin{subfigure}[c]{0.162\textwidth}
		\centering
		\includegraphics[width=1.12in]{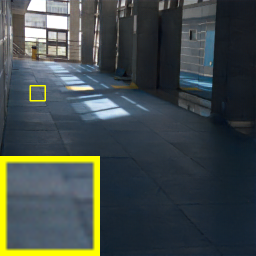}
		\vspace{-1.5em}
		\caption*{Ours}
	\end{subfigure} 
	\begin{subfigure}[c]{0.162\textwidth}
		\centering
		\includegraphics[width=1.12in]{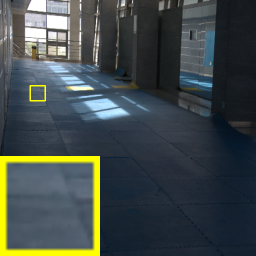}
		\vspace{-1.5em}
		\caption*{GT}
	\end{subfigure}  
	\vspace{0.2em} \\
	
	\begin{subfigure}[c]{0.162\textwidth}
		\centering
		\includegraphics[width=1.12in]{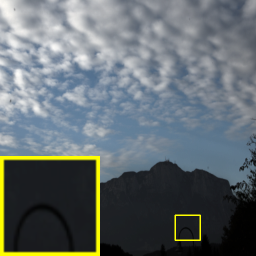}
		\vspace{-1.5em}
		\caption*{Input}
	\end{subfigure}
	\begin{subfigure}[c]{0.162\textwidth}
		\centering
		\includegraphics[width=1.12in]{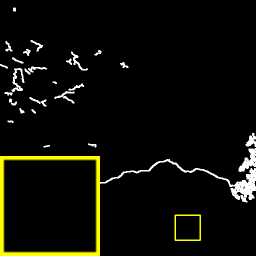}
		\vspace{-1.5em}
		\caption*{Structure of Input}
	\end{subfigure}
	\begin{subfigure}[c]{0.162\textwidth}
		\centering
		\includegraphics[width=1.12in]{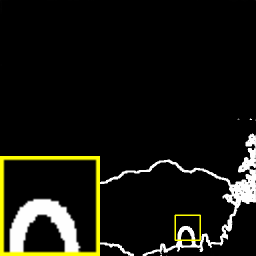}
		\vspace{-1.5em}
		\caption*{Our Structure Map}
	\end{subfigure}
	\begin{subfigure}[c]{0.162\textwidth}
		\centering
		\includegraphics[width=1.12in]{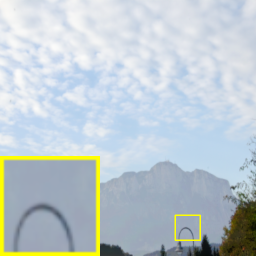}
		\vspace{-1.5em}
		\caption*{SNR~\cite{xu2022snr}}
	\end{subfigure}
	\begin{subfigure}[c]{0.162\textwidth}
		\centering
		\includegraphics[width=1.12in]{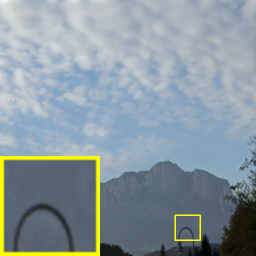}
		\vspace{-1.5em}
		\caption*{Ours}
	\end{subfigure} 
	\begin{subfigure}[c]{0.162\textwidth}
		\centering
		\includegraphics[width=1.12in]{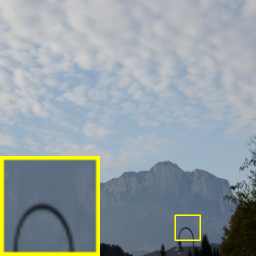}
		\vspace{-1.5em}
		\caption*{GT}
	\end{subfigure} 
	\vspace{-0.133in}
	\caption{Visual comparisons on LOL-real (top) and LOL-synthetic (bottom). ``Ours'' have less noise and clearer visibility.}
	\vspace{-0.2in}
	\label{fig:cmp_lol2r}
\end{figure*}

\subsection{Implementation Details}
We implement our framework in PyTorch~\cite{paszke2019pytorch}.
Due to the limited computation resource, our framework is trained on 4 GPUs with RTX3090.
For the loss minimization, we adopt the Adam~\cite{kingma2014adam} optimizer with momentum set to 0.9.

\vspace{-0.05in}
\subsection{Datasets}
\vspace{-0.05in}
Existing low-light enhancement datasets have different properties.
SID~\cite{chen2018learning} is a challenging dataset in both sRGB and RAW domains, and many dark areas need to be enhanced with structural guidance.
For SID, each input sample is a pair of short- and long-exposure images, and low-light images have heavy noise since they were captured in extremely dark environments.
We use the subset captured by the Sony camera of SID for experiments.
Moreover, we also adopt LOL~\cite{yang2021sparse} dataset, which is divided into LOL-real and LOL-synthetic.
Low-light images in LOL-real were collected from the real world; LOL-synthetic was created by synthesis.

\vspace{-0.05in}
\subsection{Comparison on sRGB Domain}
\vspace{-0.05in}
We compare our method with a rich collection of SOTA methods for low-light enhancement, including
SID~\cite{chen2018learning}, DeepUPE~\cite{wang2019underexposed}, KIND~\cite{zhang2019kindling}, DeepLPF~\cite{moran2020deeplpf}, FIDE~\cite{xu2020learning}, LPNet~\cite{li2020luminance}, MIR-Net~\cite{zamir2020learning}, RF~\cite{kosugi2020unpaired}, 3DLUT~\cite{zeng2020learning}, A3DLUT~\cite{wang2021real}, Band~\cite{yang2021band}, EG~\cite{jiang2021enlightengan}, Retinex~\cite{liu2021retinex}, Sparse~\cite{yang2021sparse}, UNIE~\cite{jin2022unsupervised}, LCDR~\cite{wang2022local}, LLFlow~\cite{wang2022low}, DSN~\cite{zhao2021deep}, RCTNet~\cite{kim2021representative}, UTVNet~\cite{zheng2021adaptive}, SCI~\cite{ma2022toward}, URetinex~\cite{wu2022uretinex}, and SNR~\cite{xu2022snr}.

\noindent\textbf{Quantitative analysis.} \
We adopt PSNR and SSIM~\cite{wang2004image} for evaluation.
A higher SSIM means more high-frequency details and structures in results.
Tables~\ref{comparison3} and \ref{comparison4} show the comparisons on LOL-real and LOL-synthetic.
Our method surpasses all the baselines, and the improvement of SSIM is very evident thanks to the successful structure modeling and guidance.
These numbers are obtained either from the respective papers or by running the respective public code.
Table~\ref{comparison5} shows the comparison on SID.
It can be seen that although previous SOTA approaches, e.g., SNR~\cite{xu2022snr}, can achieve high PSNR, their corresponding SSIM values are not satisfying. Our framework again has the best performance over others and has obvious superiority to others by large margins on SSIM.

\begin{figure*}[t]
	\centering
	\begin{subfigure}[c]{0.162\textwidth}
		\centering
		\includegraphics[width=1.12in]{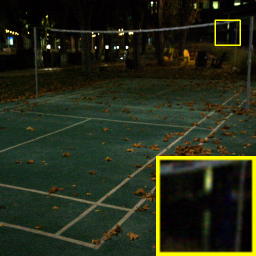}
		\vspace{-1.5em}
		\caption*{Input}
	\end{subfigure}
	\begin{subfigure}[c]{0.162\textwidth}
		\centering
		\includegraphics[width=1.12in]{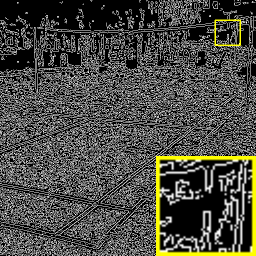}
		\vspace{-1.5em}
		\caption*{Structure of Input}
	\end{subfigure}
	\begin{subfigure}[c]{0.162\textwidth}
		\centering
		\includegraphics[width=1.12in]{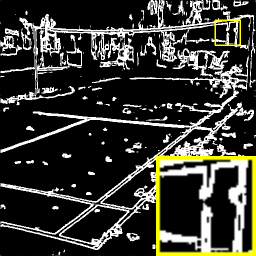}
		\vspace{-1.5em}
		\caption*{Our Structure Map}
	\end{subfigure}
	\begin{subfigure}[c]{0.162\textwidth}
		\centering
		\includegraphics[width=1.12in]{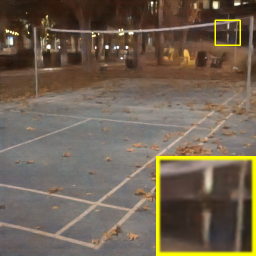}
		\vspace{-1.5em}
		\caption*{SNR~\cite{xu2022snr}}
	\end{subfigure}
	\begin{subfigure}[c]{0.162\textwidth}
		\centering
		\includegraphics[width=1.12in]{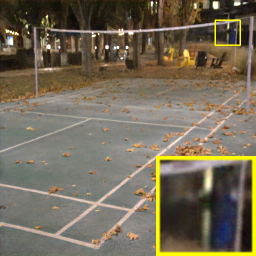}
		\vspace{-1.5em}
		\caption*{Ours}
	\end{subfigure} 
	\begin{subfigure}[c]{0.162\textwidth}
		\centering
		\includegraphics[width=1.12in]{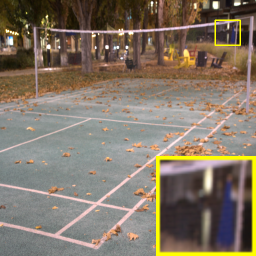}
		\vspace{-1.5em}
		\caption*{GT}
	\end{subfigure}  \vspace{0.2em} \\
	
	\begin{subfigure}[c]{0.162\textwidth}
		\centering
		\includegraphics[width=1.12in]{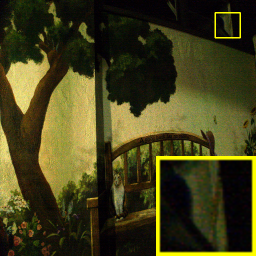}
		\vspace{-1.5em}
		\caption*{Input}
	\end{subfigure}
	\begin{subfigure}[c]{0.162\textwidth}
		\centering
		\includegraphics[width=1.12in]{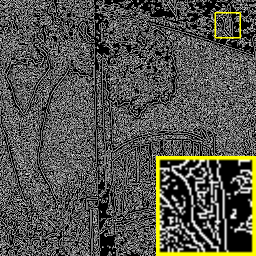}
		\vspace{-1.5em}
		\caption*{Structure of Input}
	\end{subfigure}
	\begin{subfigure}[c]{0.162\textwidth}
		\centering
		\includegraphics[width=1.12in]{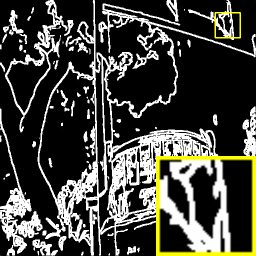}
		\vspace{-1.5em}
		\caption*{Our Structure Map}
	\end{subfigure}
	\begin{subfigure}[c]{0.162\textwidth}
		\centering
		\includegraphics[width=1.12in]{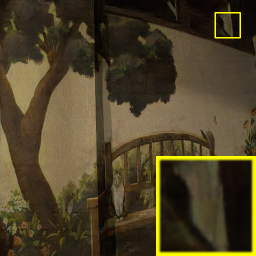}
		\vspace{-1.5em}
		\caption*{SNR~\cite{xu2022snr}}
	\end{subfigure}
	\begin{subfigure}[c]{0.162\textwidth}
		\centering
		\includegraphics[width=1.12in]{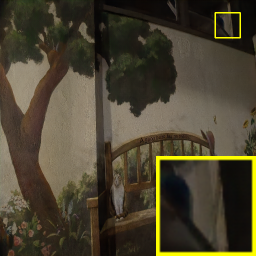}
		\vspace{-1.5em}
		\caption*{Ours}
	\end{subfigure} 
	\begin{subfigure}[c]{0.162\textwidth}
		\centering
		\includegraphics[width=1.12in]{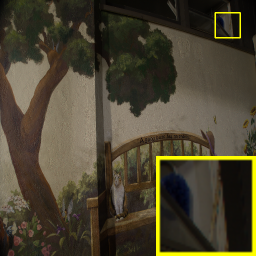}
		\vspace{-1.5em}
		\caption*{GT}
	\end{subfigure} 
	\vspace{-0.133in}
	\caption{Visual comparisons on SID's sRGB (top) and RAW (bottom) domains. ``Ours'' have less noise and clearer visual details.}
	\vspace{-0.2in}
	\label{fig:cmp_sid}
\end{figure*}

\noindent\textbf{Qualitative analysis.} \
We present visual samples in Fig.~\ref{fig:cmp_lol2r} to compare our method with the baseline that achieves the best performance on LOL-real and LOL-synthetic.
Our result shows better visual quality with higher contrast, more accurate and clear details, more natural color consistency and brightness. 
Fig.~\ref{fig:cmp_sid} also shows visual comparisons on SID.
While the input images in SID have evident noise and weak illumination, our method can produce more realistic results with fewer artifacts than others.

\subsection{Comparison on RAW Domain}
We compare our method with existing low-light image enhancement methods that are designed for RAW inputs.
The baselines include CAN~\cite{chen2017fast}, DeepUPE~\cite{wang2019underexposed}, SID~\cite{chen2018learning}, EEMEFN~\cite{zhu2020eemefn}, LLPackNet~\cite{lamba2020towards}, FIDE~\cite{xu2020learning}, DID~\cite{maharjan2019improving}, SGN~\cite{gu2019self}, DCE~\cite{Zero-DCE}, RED~\cite{lamba2021restoring}, and ABF~\cite{dong2022abandoning}.
Compared with the enhancement on the sRGB domain, the RAW inputs have more information, leading to the enhancement with higher PSNR and SSIM.

The quantitative results are shown in Table~\ref{comparison6}. Our method still obtains the SOTA performance on the RAW domain in terms of both PSNR and SSIM. 
Our method results in the best SSIM that characterize the richness and sharpness of the images, since our approach formulates the effective structure modeling explicitly.
The qualitative results in Fig.~\ref{fig:cmp_sid} also support the superiority of our framework in the RAW domain. As shown in the figure, these results also manifest that our method effectively enhances the image lightness and reveals details, while suppressing artifacts.

\begin{table}[t]
	\centering
	\resizebox{1.0\linewidth}{!}{
		\begin{tabular}{c|ccccc}
			\toprule[1pt]
			Methods & DeepUPE~\cite{wang2019underexposed} &SID~\cite{chen2018learning}& EEMEFN~\cite{zhu2020eemefn}&DCE~\cite{Zero-DCE}\\
			PSNR &29.13&28.88&29.60& 26.53 \\
			SSIM&0.792&0.787&0.795&0.730 \\
			\hline
			Methods &LLPackNet~\cite{lamba2020towards} &FIDE~\cite{xu2020learning}&DID~\cite{maharjan2019improving}&SGN~\cite{gu2019self} \\
			PSNR&27.83 & 29.56&28.41&28.91\\
			SSIM&0.750 &0.799&0.780&0.789\\
			\hline
			Methods  &RED~\cite{lamba2021restoring}&ABF~\cite{dong2022abandoning}&SNR~\cite{xu2022snr}&Ours \\
			PSNR&28.66&29.65& 29.75&\textbf{30.17}\\
			SSIM&0.790&0.797&0.812 &\textbf{0.834}\\
			\bottomrule[1pt]
	\end{tabular}}
	\vspace{-0.15in}
	\caption{Quantitative comparison on SID's RAW domain.}\vspace{-0.25in}
	\label{comparison6}
\end{table}

\vspace{-0.05in}
\subsection{Ablation Study}
\label{sec:ab}
We consider the following ablation settings by removing different components from our framework individually.
\vspace{-0.05in}
\begin{enumerate}
	\item ``Ours w/o $\mathcal{A}$": remove the module of $\mathcal{A}$, only the input image and the structure map are set as the input of $\mathcal{E}$.
	\item ``Ours w/o $\mathcal{S}$": delete the structural modeling module $\mathcal{S}$, and the framework is changed to the structure of two concatenated networks for appearance modeling.
	\item ``Ours w/o $\mathcal{F}$": replace the SAFE with the traditional encoder for the StyleGAN~\cite{yang2021gan}.
	\item ``Ours w/o $\mathcal{G}$": remove structure-guided feature synthesis in $\mathcal{E}$, and set the output of $\mathcal{S}$ as the input of $\mathcal{E}$. 
	\item ``Ours w/o S.G.": use the SOTA encoder-decoder-based edge prediction network~\cite{pu2022edter} to implement $\mathcal{S}$.
	\item ``Ours w/o GAN": train $\mathcal{S}$ without GAN loss.
\end{enumerate}
We performed ablation studies on all three datasets on the sRGB domain.
Table~\ref{abla} summarizes the results.
Compared with all ablation settings, our full setting yields higher PSNR and SSIM.
Comparing ``Ours w/o $\mathcal{A}$" with ``Ours" shows the necessity of conducting both the appearance and structure modeling in our framework. And the comparison between ``Ours w/o $\mathcal{S}$" and ``Ours" demonstrates the effectiveness of the structure guidance to enhance appearance modeling.
The results also show effects of ``SAFE'' (``Ours w/o $\mathcal{F}$" vs. ``Ours"), ``structure-guided feature synthesis" (``Ours w/o $G$" vs. ``Ours"), and the GAN loss in structure modeling (``Ours w/o GAN" vs. ``Ours").
Moreover, comparing ``Ours w/o S.G." with ``Ours" proves the superiority of our $\mathcal{S}$ over existing encoder-decoder-based edge detection networks on low-light images.

Furthermore, to demonstrate the robustness of our framework, we conduct the evaluation setting where extra noise (the Gaussian distribution with mean as 0, variance ranges from 30 to 50) is added in the input image. The results are displayed in Table~\ref{abla} as ``Ours with noise", which is close to the situation without noise, showing the robustness of our framework towards perturbations.

\begin{table}[t]
	\centering
	\resizebox{1.0\linewidth}{!}{
		\begin{tabular}{l|cc|cc|cc}
			\hline
			&\multicolumn{2}{c|}{LOL-real}&\multicolumn{2}{c|}{LOL-synthetic}& \multicolumn{2}{c}{SID}\\
			\hline
			Methods & PSNR & SSIM& PSNR & SSIM& PSNR & SSIM \\
			\hline
			Ours w/o $\mathcal{A}$
			&20.17&0.801&23.59&0.879&22.59&0.639\\
			Ours w/o $\mathcal{S}$
			&18.14&0.773&21.20&0.881&20.47&0.623\\
			Ours w/o $\mathcal{F}$
			&20.21&0.812&23.05&0.888&22.35&0.635\\
			Ours w/o $\mathcal{G}$
			&19.39&0.784&21.71&0.868&21.15&0.629\\
			Ours w/o S.G.
			&20.73 &0.820 & 23.30&0.898 &22.50 &0.632\\
			Ours w/o GAN
			&21.28&0.812&23.17&0.883&22.14&0.642\\
			\hline
			Results with $\mathcal{A}$ &18.99 &0.715 &21.76&0.863&19.34&0.556\\
			Ours with noise
			&24.15&0.832&24.07&0.880&22.86&0.648\\
			\hline
			Ours &\textbf{24.62} &\textbf{0.867} &\textbf{25.62} &\textbf{0.905} &\textbf{23.18} &\textbf{0.664}\\
			\hline
	\end{tabular}}
	\vspace{-0.15in}
	\caption{Results of the ablation study in Sec.~\ref{sec:ab}.}
	\label{abla}
	\vspace{-0.1in}
\end{table}

\begin{table}[t]
	\centering
	\resizebox{1.0\linewidth}{!}{
		\begin{tabular}{l|cc|cc|cc}
			\hline
			&\multicolumn{2}{c|}{LOL-real}&\multicolumn{2}{c|}{LOL-synthetic}& \multicolumn{2}{c}{SID}\\
			\hline
			Methods & CE & $L_2$& CE & $L_2$&CE & $L_2$ \\
			\hline
			Ours w/o GAN
			&0.2581&0.3650&0.2144&0.3936&0.5335&0.5034\\
			Ours w/o S.G.
			&0.2923&0.3805&0.2133&0.3833&0.5035&0.5405\\
			Ours w/o $\mathcal{F}$
			&0.3070&0.3553&0.2795&0.3675&0.5351&0.4905\\
			Ours &\textbf{0.2130} &\textbf{0.3042} &\textbf{0.2072} &\textbf{0.3032} &\textbf{0.4352} &\textbf{0.4541}\\
			\hline
	\end{tabular}}
	\vspace{-0.15in}
	\caption{Results of the ablation study in Sec.~\ref{sec:ab2}.}
	\label{abla2}
	\vspace{-0.1in}
\end{table}

\begin{figure}[t]
	\centering
	\small
	\begin{subfigure}[c]{0.073\textwidth}
		\centering
		\includegraphics[width=0.52in]{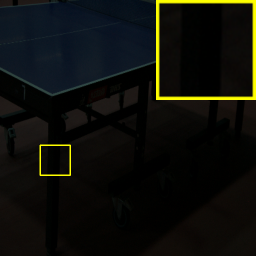}
		\vspace{-1.5em}
		\caption*{Input}
	\end{subfigure}
	\begin{subfigure}[c]{0.073\textwidth}
		\centering
		\includegraphics[width=0.52in]{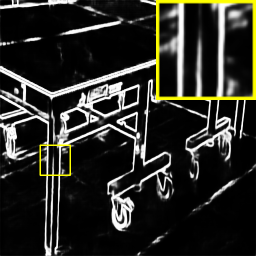}
		\vspace{-1.5em}
		\caption*{w/o GAN}
	\end{subfigure}
	\begin{subfigure}[c]{0.073\textwidth}
		\centering
		\includegraphics[width=0.52in]{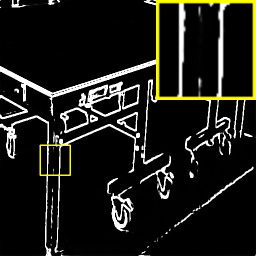}
		\vspace{-1.5em}
		\caption*{w/o S.G.}
	\end{subfigure}
	\begin{subfigure}[c]{0.073\textwidth}
		\centering
		\includegraphics[width=0.52in]{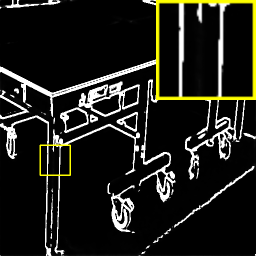}
		\vspace{-1.5em}
		\caption*{w/o $\mathcal{F}$}
	\end{subfigure} 
	\begin{subfigure}[c]{0.073\textwidth}
		\centering
		\includegraphics[width=0.52in]{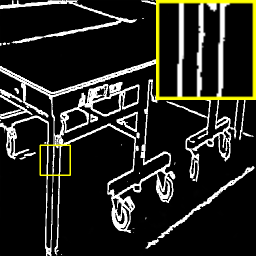}
		\vspace{-1.5em}
		\caption*{Ours}
	\end{subfigure}  
	\begin{subfigure}[c]{0.073\textwidth}
		\centering
		\includegraphics[width=0.52in]{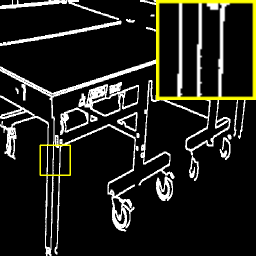}
		\vspace{-1.5em}
		\caption*{GT}
	\end{subfigure}  
	\vspace{0.2em} \\
	
	\begin{subfigure}[c]{0.073\textwidth}
		\centering
		\includegraphics[width=0.52in]{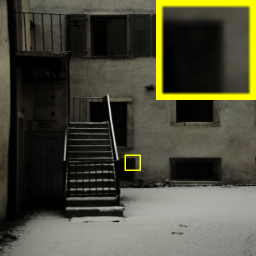}
		\vspace{-1.5em}
		\caption*{Input}
	\end{subfigure}
	\begin{subfigure}[c]{0.073\textwidth}
		\centering
		\includegraphics[width=0.52in]{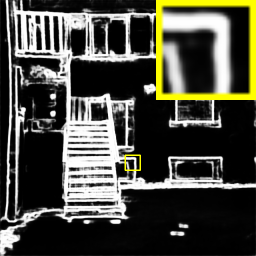}
		\vspace{-1.5em}
		\caption*{w/o GAN}
	\end{subfigure}
	\begin{subfigure}[c]{0.073\textwidth}
		\centering
		\includegraphics[width=0.52in]{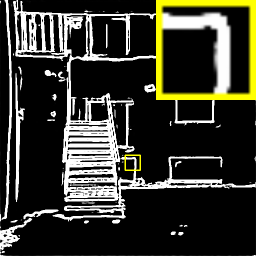}
		\vspace{-1.5em}
		\caption*{w/o S.G.}
	\end{subfigure}
	\begin{subfigure}[c]{0.073\textwidth}
		\centering
		\includegraphics[width=0.52in]{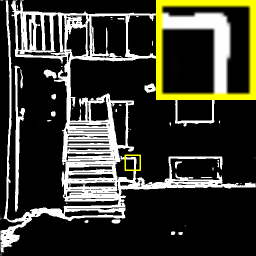}
		\vspace{-1.5em}
		\caption*{w/o $\mathcal{F}$}
	\end{subfigure} 
	\begin{subfigure}[c]{0.073\textwidth}
		\centering
		\includegraphics[width=0.52in]{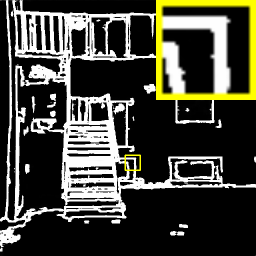}
		\vspace{-1.5em}
		\caption*{Ours}
	\end{subfigure}  
	\begin{subfigure}[c]{0.073\textwidth}
		\centering
		\includegraphics[width=0.52in]{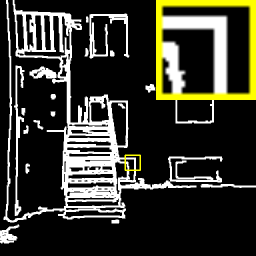}
		\vspace{-1.5em}
		\caption*{GT}
	\end{subfigure}  
	\vspace{0.2em} \\
	
	\begin{subfigure}[c]{0.073\textwidth}
		\centering
		\includegraphics[width=0.52in]{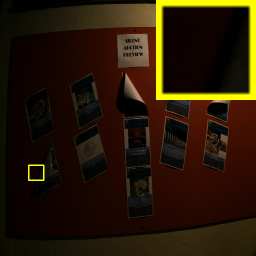}
		\vspace{-1.5em}
		\caption*{Input}
	\end{subfigure}
	\begin{subfigure}[c]{0.073\textwidth}
		\centering
		\includegraphics[width=0.52in]{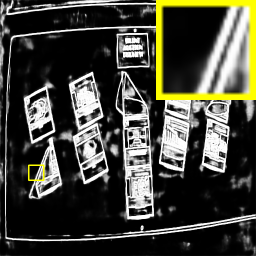}
		\vspace{-1.5em}
		\caption*{w/o GAN}
	\end{subfigure}
	\begin{subfigure}[c]{0.073\textwidth}
		\centering
		\includegraphics[width=0.52in]{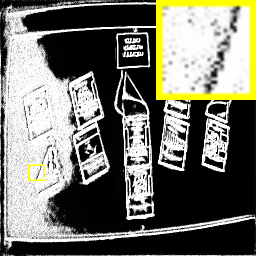}
		\vspace{-1.5em}
		\caption*{w/o S.G.}
	\end{subfigure}
	\begin{subfigure}[c]{0.073\textwidth}
		\centering
		\includegraphics[width=0.52in]{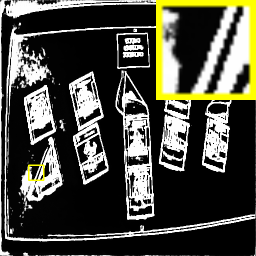}
		\vspace{-1.5em}
		\caption*{w/o $\mathcal{F}$}
	\end{subfigure} 
	\begin{subfigure}[c]{0.073\textwidth}
		\centering
		\includegraphics[width=0.52in]{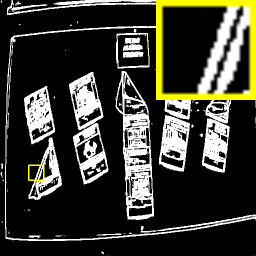}
		\vspace{-1.5em}
		\caption*{Ours}
	\end{subfigure}  
	\begin{subfigure}[c]{0.073\textwidth}
		\centering
		\includegraphics[width=0.52in]{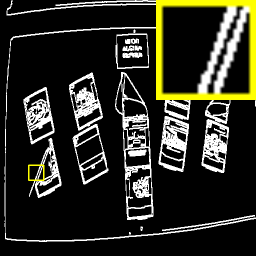}
		\vspace{-1.5em}
		\caption*{GT}
	\end{subfigure}  \\
	\vspace{-0.133in}
	\caption{Visual comparisons for structure modeling on LOL-real, LOL-synthetic, and SID datasets (top to bottom).}
	\vspace{-0.2in}
	\label{fig:cmp_structure}
\end{figure}

\begin{figure*}[t]
	\begin{center} 
		\includegraphics[width=1.0\linewidth]{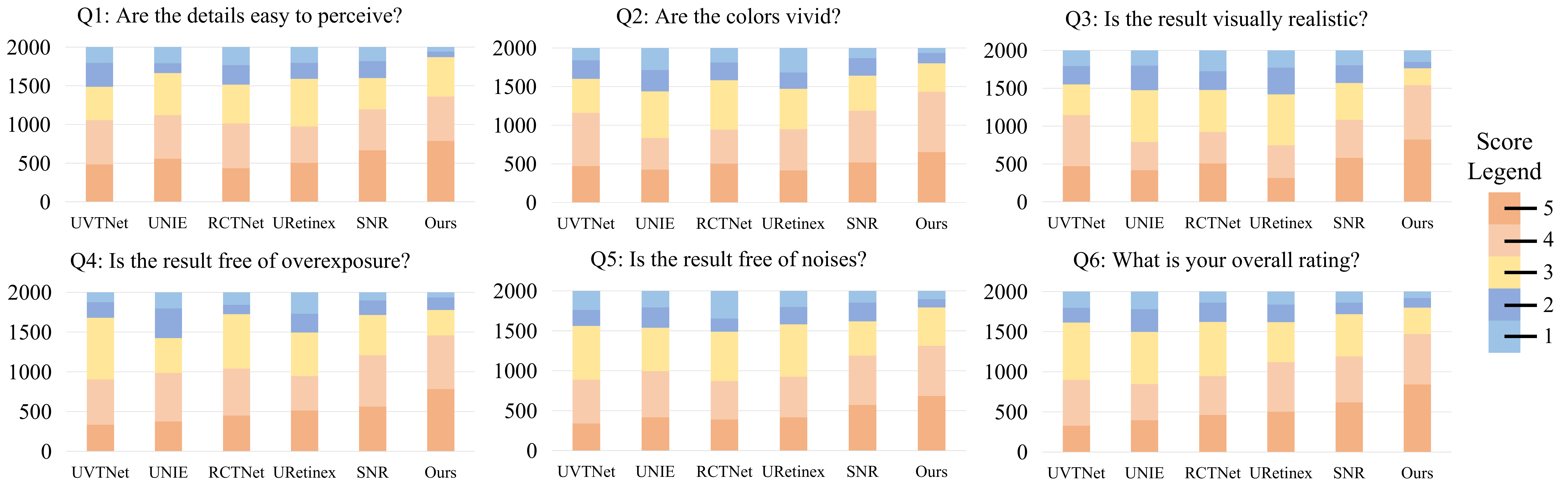}
	\end{center}
	\vspace{-0.2in}
	\caption{The results of the user study. The ordinate axis denotes the rating frequency received from all participants.}
	\label{fig:user_study}
	\vspace{-0.2in}
\end{figure*}

\subsection{Evaluation for Structural Modeling}
\label{sec:ab2}
In this section, we conduct ablation studies for the structure modeling.
Since we set the edge map of the normal-light image as the ground truth, we use the cross-entropy (CE) and the $L_2$ distance between the prediction and the ground truth as the metric.
We compare the ablation settings that also employ the structure modeling, i.e., ``Ours w/o GAN", ``Ours w/o S.G.", and ``Ours w/o $\mathcal{F}$".
The results are reported in Table~\ref{abla2}. Compared with all ablation settings, our full setting achieves the most accurate structure modeling from the low-light images.
By comparing ``Ours w/o $\mathcal{F}$" and ``Ours w/o S.G." with ``Ours", we can demonstrate the superiority of our $S$ in the structure modeling to other alternatives. By comparing ``Ours" with ``Ours w/o GAN", we show the effects of the GAN loss in the structure modeling from low-light images.
The visual comparisons are also provided in Fig.~\ref{fig:cmp_structure}. We can see that the edge predictions' uncertainty will be increased without the GAN loss, causing the artifacts in the edge prediction.
When compared with the results of ``Ours w/o $\mathcal{F}$" and ``Ours w/o S.G.", our results have more edge details and fewer artifacts.

\vspace{-0.05in}
\subsection{User Study}
\vspace{-0.05in}
We conduct a large-scale user study with 100 participants for human subjective evaluation, and the five strongest baselines are chosen by mean PSNR on SID and LOL. All methods are trained on SID, and we evaluate their performances on the real-captured low-light photos (a total of 50 images) from an iPhone 8.
The capturing environments are various, including indoor and outdoor scenes.
Following the settings in~\cite{xu2022snr}, the evaluation is completed via user ratings on the six questions shown in Fig.~\ref{fig:user_study}, where scores range from 1 (worst) to 5 (best). 
Fig.~\ref{fig:user_study} exhibits the rating distributions of different methods. We can see that our framework receives more high and fewer low ratings, showing the superior perception of our results.

\begin{table}[t]
	\centering
	\resizebox{1.0\linewidth}{!}{
		\begin{tabular}{l|cc|cc|cc}
			\hline
			&\multicolumn{2}{c|}{LOL-real}&\multicolumn{2}{c|}{LOL-synthetic}& \multicolumn{2}{c}{SID}\\
			\hline
			Methods & PSNR & SSIM& PSNR & SSIM& PSNR & SSIM \\
			\hline
			Ours with E.D.
			&25.57&0.885&26.48&0.951&24.43&0.707\\
			Ours with Seg.
			&24.18&0.846&25.27&0.903&23.10&0.659\\
			Ours with Dep.
			&23.89&0.830&24.91&0.902&22.95&0.640\\
			\hline
			Ours &24.62 &0.867 &25.62 &0.905 &23.18 &0.664\\
			\hline
	\end{tabular}}
	\vspace{-0.1in}
	\caption{Results of experiments in Sec.~\ref{sec:smmd} and \ref{sec:sr}.}
	\label{abla3}
	\vspace{-0.2in}
\end{table}

\vspace{-0.05in}
\subsection{Structure Modeling with More Data}
\label{sec:smmd}
\vspace{-0.05in}
All the above results (ours and baselines) are obtained from the networks trained without extra data.
The collection of paired low-light and normal-light data, i.e., $I$ and $\bar{I}$, is difficult and expensive, while the paired data to supervise the structure modeling $\mathcal{S}$ is trivial since the ground truth $\bar{I}_s$ can be obtained with the edge detection algorithm from $\bar{I}$.
Therefore, we can easily obtain extra data pair for $\mathcal{S}$ from another dataset containing normal-light data. Following \cite{yang2020fidelity}, we adopt AVA dataset~\cite{murray2012ava} to acquire the data pair to finetune $\mathcal{S}$. The result can be improved again by synthesizing better structure for very dark areas, as shown in Table~\ref{abla3} (Ours with E.D.).

\vspace{-0.05in}
\subsection{Ours with Other Structural Representations}
\label{sec:sr}
\vspace{-0.05in}
In this work, we adopt the edge for the structure modeling since the edge is a general structural representation for various scenes. Our framework is also applicable to other structural representations, e.g., the segmentation and depth maps.
We adopt the DPT model~\cite{ranftl2021vision}, which is trained on ADE20K~\cite{zhou2017scene} and MIX 6~\cite{ranftl2021vision}, to extract the normal-light data's segmentation maps and depth maps. And they are employed as the ground truth to train $\mathcal{S}$. 
As shown in Table~\ref{abla3}, all these results (``Ours with Seg." and ``Ours with Dep.") are better than the baselines in Tables~\ref{comparison3},~\ref{comparison4},~\ref{comparison5}, demonstrating the effectiveness of our structure modeling and guidance strategy.

\vspace{-0.05in}
\section{Conclusion}
\vspace{-0.05in}
In this paper, we propose a new framework to conduct structure modeling and employ the restored structure maps to enhance the appearance modeling results. Different from existing methods, the structure modeling in our framework is undertaken via a modified generative model with structural features and is trained with the GAN loss. We further design a novel structure-guided enhancement module for appearance enhancement with structure-guided feature synthesis layers. Extensive experiments on sRGB and RAW domains demonstrate the effectiveness of our framework.

\noindent\textbf{Limitation.} The structure generation with our model in extremely dark areas (with almost no information) could have artifacts when only trained with existing low-light datasets. This issue can be alleviated by training with large-scale data as indicated in Sec.~\ref{sec:smmd}, requiring more computation resources and a long training time.

\noindent\textbf{Acknowledgments.}
This work is partially supported by Shenzhen Science and Technology Program KQTD20210811090149095.

{\small
	\bibliographystyle{ieee_fullname}
	\bibliography{egbib}
}

\end{document}